%% file: main.tex
\documentclass[10pt,letterpaper]{article}

\usepackage{cogsci}

\input{config}
\usepackage{pslatex}
\usepackage{apacite}
\usepackage{float} 

\title{Individual vs. Joint Perception:\\a Pragmatic Model of Pointing as Communicative Smithian Helping}
 
\author{
    \begin{tabular}{c c c c}
        \bf Kaiwen Jiang$^{1}$ \quad\quad & \bf Stephanie Stacy$^{1}$  \quad\quad & \bf Chuyu Wei$^{3}$ \quad\quad & Adelpha Chan$^{4}$ \\
        \normalfont kaiwenj@g.ucla.edu \quad\quad & \normalfont stephaniestacy@g.ucla.edu \quad\quad & \normalfont 
        chuyuwei@g.ucla.edu
        \quad\quad & \normalfont 
        adelchan07@g.ucla.edu
    \end{tabular}
    \\\vspace{-9pt}\\
    \begin{tabular}{c c c}
        \bf Federico Rossano$^{5}$ & \bf Yixin Zhu$^{1}$ \quad\quad & \bf Tao Gao$^{1,2}$\\
        \normalfont frossano@ucsd.edu \quad\quad & \normalfont yixin.zhu@ucla.edu & \normalfont tao.gao@stat.ucla.edu
    \end{tabular}
    \\\vspace{-9pt}\\
    \begin{tabular}{c c c}
        $^1$ Department of Statistics, UCLA & $^2$ Department of Communication, UCLA & $^3$ Department of Psychology, UCLA
    \end{tabular}
    \\\vspace{-11pt}\\
    \begin{tabular}{c c}
        $^4$ Department of Linguistics, UCLA & $^5$ Department of Cognitive Science, UCSD
    \end{tabular}
}

\begin{document}

\maketitle

\begin{abstract}
The simple gesture of pointing can greatly augment one's ability to comprehend states of the world based on observations. It triggers additional inferences relevant to one's task at hand. We model an agent's update to its belief of the world based on individual observations using a \acf{pomdp}, a mainstream \acf{ai} model of how to act rationally according to beliefs formed through observation. On top of that, we model pointing as a communicative act between agents who have a mutual understanding that the pointed observation must be relevant and interpretable. Our model measures ``relevance'' by defining a \acf{svi} as the utility improvement of the \ac{pomdp} agent before and after receiving the pointing. We model that agents calculate \ac{svi} by using the cognitive theory of Smithian helping as a principle of coordinating separate beliefs for action prediction and action evaluation. We then import \ac{svi} into \acf{rsa} as the utility function of an utterance. These lead us to a pragmatic model of pointing allowing for contextually flexible interpretations. 
We demonstrate the power of our Smithian pointing model by extending the Wumpus world, a classic \ac{ai} task where a hunter hunts a monster with only partial observability of the world. We add another agent as a guide who can only help by marking an observation already perceived by the hunter with a pointing or not, without providing new observations or offering any instrumental help. Our results show that this severely limited and overloaded communication nevertheless significantly improves the hunters' performance. The advantage of pointing is indeed due to a computation of relevance based on Smithian helping, as it disappears completely when the task is too difficult or too easy for the guide to help.

\textbf{Keywords:} 
pointing; pragmatics; joint attention; Smithian helping; cooperation; rational speech act
\end{abstract}

\section{Introduction}

Like all animals, we understand the world by collecting observations through our individual perception, which we call ``individually perceived observations.'' Being social creatures, however, we also get observations that are pointed out to us by others. When someone points, the addressee knows that they and the person who initiated the pointing gesture both get an observation, propagating a belief update based on the observation. We call this a ``jointly perceived'' observation. In this modeling paper, we use the term ``observation'' to refer to the raw sensory input as in the field of \acf{ai} \cite{kaelbling1998planning}. We use the term ``perception'' to refer to the inference of the most likely world state that generates the observation following the Bayesian perspective of perception \cite{knill1996perception}. This paper aims to demonstrate from a computational perspective that the joint perception enabled by pointing is more potent than the individual perception of the same observation. We aim to prove that insights of paternalistic helping from developmental psychology can inspire the development of socially capable \ac{ai} systems. Also, our mathematical modeling grounded in \ac{ai} algorithms can scaffold future cognitive science researches on Smithian (paternalistic) helping.

\subsection{Pointing Gesture in Human Communication}

\begin{quote}
    \small
    ``Point to a piece of paper. And now point to its shape---now to its color---now to its number. \ldots How did you do it?''\\
    \phantom{}\hfill{}---\citeA{wittgenstein2001philosophical} 
\end{quote}

Until Wittgenstein called attention to its underlying complexity, pointing had generally been perceived as an intuitive, unremarkable communicative gesture. It is among the most conspicuous and common forms of human communication. Children point to help adults retrieve objects they were looking for, foraging partners point out the potential locations of food to help each other, and customers point to their empty glasses to request assistance from the server. Pointing is among the first communicative gestures human infants learn to use \cite{butterworth2013development}: at as young as the age of one, infants use pointing to communicate information \cite{liszkowski2006twelveand}. Although pointing is pervasive in everyday human life, it is rarely observed in wild animals. Human-raised great apes can produce pointing-like gestures to invite humans to cooperate with them in obtaining food \cite{leavens1998intentional}, yet pointing in great apes lacks the cooperative properties of human pointing: apes are not bothered when the partner is distracted or non-responding \cite{van2014differences}. Chimpanzees' failure to achieve a deep understanding of pointing suggests that human pointing may reveal surprising intricacy of human communication.

The same pointing act can be interpreted differently in various contexts, which reveals two properties of pointing. First, pointing is \textbf{overloaded}. As \citeA{wittgenstein2001philosophical} pointed out, the same pointing gesture has many interpretations, making the referent of pointing ambiguous when considered in isolation. Second, pointing is \textbf{indirect}; a big gap can exist between the referent and the meaning of the pointing. The receiver must infer what to do with the referent beyond looking at it. It has been shown that by following an adult's pointing to a toy, young children could adaptively decide what to do with the toy, put it away or examine it, based on the context of the pointing \cite{liebal2011young}.

The overloadedness and indirectness of pointing enable it to express manifold meanings with the same observation, no new observations provided. The meaning of the pointing can be interpreted depending on the context. This makes pointing a powerful communicative act that can significantly facilitate human cooperation. For example, when two hunters walk in a forest together, the young hunter perceives a broken stick on the ground but does not think it is relevant to the hunt. Just the moment he is about to move on, the experienced hunter grabs his attention and points to the same broken stick he has already perceived. The young hunter immediately realizes that the broken stick is a trace of their prey. This example highlights that joint perception enabled by pointing can evoke richer inferences than the individual perception of the same observation \cite{sperber1986relevance}.

As in the above example, a crucial function of pointing is helping. Crucially, it is a particular type of helping with two unique characteristics. First, the helper is in a position to help because her belief is closer to reality, not because she has any physical advantage. Therefore, pointing must involve diverging beliefs in which the helper knows better how to improve the helpee's well-being. This type of helping is called paternalistic helping or Smithian helping in developmental psychology \cite{martin2016you}, which we will introduce later. Second, unlike instrumental actions, pointing does not change the physical states at all. Its only function is to change the helpee's mind. Therefore, models of instrumental helping would fail to apply \cite{ullman2009help}. Instead, we argue that it should be understood as an ``utterance'' in \acf{rsa}, a pragmatic model of language which also views language as cooperative \cite{frank2012predicting,goodman2016pragmatic}. \ac{rsa} treats an utterance as an action with a utility function. The generation and interpretation of an utterance can be modeled with the principle of maximizing expected utility from decision theory. 

Due to the above two unique characteristics, we propose to model pointing as an utterance with a utility derived from Smithian helping for coordinating diverging beliefs.

\subsection{Smithian Empathy and Helping}

The concept of Smithian helping is based on Adam Smith's discussion of empathy. During his discussion, he compared two types of empathy. First, he addressed Hume's definition, which proposes that empathy is a simple resonance of other's feelings. In contrast to this conventional definition, Smith proposed that true empathy involves the coordination of mindsets between the empathizer and the agent being empathized with. To better understand the two competing views addressed in Smith's argument, one can imagine a theoretical example involving you and your friend. Both of you are backstage preparing for your friend's performance in the school talent show. While your friend is excited to perform, you dread the performance because you know that he is objectively bad at singing.

\begin{figure*}[t!]
    \centering
    \includegraphics[width=\linewidth]{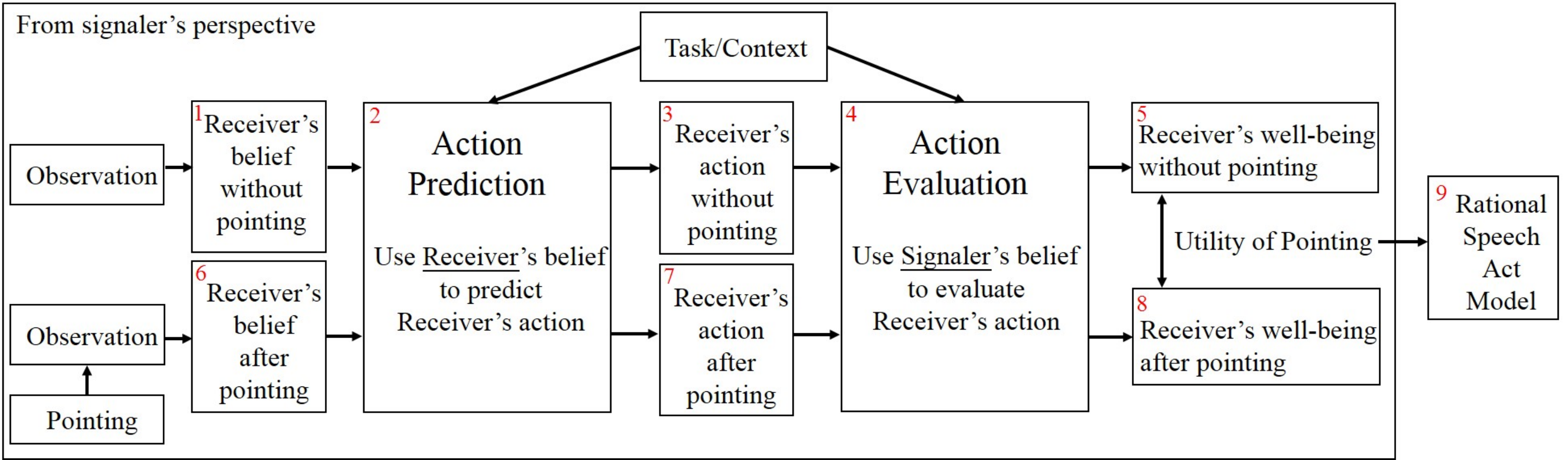}
    \caption{\textbf{Modeling pointing using Smithian coordination of beliefs}: action prediction using receiver's belief, action evaluation using signaler's belief. Numbered boxes represent key components of the model.}
    \label{fig:model}
\end{figure*}

Hume's conventional definition of empathy proposes that empathy is the \textbf{direct mirroring} of another's mindset \cite{hume2018enquiry}. In other words, the agent and subject involved in a particular act of empathy should share the same mindset. In the talent show example, you and your friend each have a different perspective in regards to the same action of your friend performing. However, according to Hume's definition, to successfully empathize with them, you must abandon your personal opinion and take on your friend's perspective.

On the other hand, Smith's own proposed definition of empathy is the act of investigating how an agent would feel if they, in their current state of consciousness, were placed into the target individual's situation \cite{smith2010theory}. In the talent show example, you and your friend each hold a distinct perspective regarding the action of your friend performing in the talent show. When executing the Smithian empathy, you maintain your own mindset when evaluating the action of interest. As a result, empathizing involves applying your own mindset to evaluate the situation of your friend performing. Because you know your friend is bad at singing, your evaluation of how you would feel in their situation leads you to be worried that your friend may embarrass himself on stage. As illustrated in the example, Smithian empathy involves \textbf{coordinating diverging mindsets} of two agents. Your act of worrying is an attempt to coordinate your friend's belief of excitement about the upcoming performance with your opposing belief that their performance will have a bad outcome. 

Smithian empathy has been explicitly extended to the well-studied phenomenon of human behavior known as paternalistic helping. Paternalistic helping involves a helper doing what she thinks to be good to the helpee, even if that is not what the helpee wants \cite{martin2016you}. This act involves the coordination of two mindsets as the helper must balance the helpee's desire with what she personally judges to be the best for the helpee. The helper acts to optimize the well-being of the helpee. In the talent show example, you believe that your friend will utterly embarrass himself if he performs. With this belief in mind, you become anxious and feel a strong urge to help your friend by convincing him not to perform. This action of stopping your friend from performing is an example of paternalistic helping as it opposes their desire to perform. You stopping your friend is driven by a sense of worry that stems from an opinion formed by applying your unique perspective to your friend's situation. On the surface, paternalistic seems deeply related to the ability to reason about other's beliefs, but it is arguably more complicated. In the famous false belief task \cite{wimmer1983beliefs}, the child only needs to select one belief to predict other's actions. In paternalistic helping, one not only needs to predict action with others' beliefs but also evaluate the actions with their own belief.

Paternalistic helping is a behavior prevalent in children that has been well studied. It has been shown that children will override a request if they recognize that following the request might harm the requester. When children interact with another child who expressed a preference for chocolate over fruit snacks but would be sick after eating chocolate, most override the request for chocolate and offer the fruit snacks \cite{martin2016you}. In addition, when the wrong tool is requested, children offer the tool they believe the requester needs, not the one they are asked for \cite{hepach2020chimpanzees}. These studies provide a solid theoretical foundation on how to coordinate two beliefs in helping. We build upon these insights and focus on cases in which helping behavior is executed to provide information through communication, as in pointing.

\section{Modeling Individual vs. Joint Perception}

Our modeling study is directly inspired by the perception of the broken stick in the hunting example. The ultimate goal here is to demonstrate that an intelligent agent with joint perception enabled by pointing can outperform an agent only with individual perception in a hunting task.

We start by outlining the model of an intelligent agent that acts based on individual perception. On top of that, we formulate a model of pointing for joint perception. The components of the models are shown in \cref{fig:model}.

\subsection{Modeling Agent with Individual Perception Using \acs{pomdp}}

For modeling an intelligent agent with individual perception, we use the \acf{pomdp} \cite{kaelbling1998planning}. \ac{pomdp} provides a generic formulation of how an agent takes rational actions in an uncertain environment with only limited observations. It models two processes in the agent-environment interaction.

The first process is how an agent updates its belief of the world state $s$ with observations from the environment using Bayesian inference (resulting in Box 1 in \cref{fig:model}). A belief is defined as a probabilistic distribution over the set of possible states:
\begin{equation}
    b\left(s\right)=P\left(s| b\right).
\end{equation}
This definition of belief is used in both \ac{ai} \cite{kaelbling1998planning} and cognitive modeling \cite{baker2011bayesian}. The agent updates its belief when it gets an observation $o$ after taking action $a$. Let $b'$ be the updated belief and $s'$ the next state after taking $a$, we have
\begin{equation}
    \begin{aligned}
        b'\left(s'\right)&=P\left(s'\middle| b'\right)=P\left(s'| b,\ o\right)\\
&\propto P\left(o| s',a\right)\sum_{s\in\mathcal{S}}{P\left(s'|a,s\right)P\left(s| b\right)}.
    \end{aligned}
\end{equation}
This belief update only involves individual perception because the agent treats the observation only as being generated by its own interaction with the environment. It does not view it as a referent of any communicative intention as in \citeA{grice1975logic}, even with the presence of a second agent.

The second process is the decision model of how an agent takes rational actions based on its belief. Planning over beliefs involves the calculation of expected utility of each action. The conventional expected utility of an action $a$ can be defined as the expectation of utility of the outcomes $s'$ of the action \cite{russell2010artificial}:
\begin{equation}
    \E U(a|b)=\int_{s'\in\mathcal{S}} U(s')P(s'|a, b),
    \label{equ:expectation_utility}
\end{equation}
where the probability of the outcomes can be calculated with a transition model:
\begin{equation}
    P\left(s'| a, b\right)=\int_{s\in \mathcal{S}}{P\left(s'| s, a\right)P\left(s|b\right)}.
\end{equation}

The agent then selects an action expected to maximize its utility: 
\begin{equation}
    a^*=\mathrm{argmax}_a \E U\left(a\right).
\end{equation}
In a \ac{pomdp}, the actions of a rational agent is determined by its belief. With knowledge of an agent's belief, the agent's actions (Box 3 in \cref{fig:model}) can be predicted. This action prediction process is represented in \cref{fig:model} as Box 2. 

Here we only outline the general principle of utility calculation in belief space. Planning rational actions in belief space to optimize long-term accumulated rewards is a challenging problem. In our study, we use the point-based value iteration (PBVI) algorithm \cite{pineau2003point} as the solver for \ac{pomdp}.

\subsection{Modeling Agent with Joint Perception: Smithian Pointing}

For an individual agent, the expected utility of action, belief, and the value of information provided by an observation can be calculated based on the agent's individual belief. We refer to them as the ``conventional'' utilities, which can be easily derived based on the classic theories \cite{russell2010artificial}. Our Smithian model of pointing augments these conventional utilities to reflect the coordination of two beliefs in Smithian helping: The signaler takes on the role of helper, while the receiver is the helpee. Specifically, the signaler uses the receiver's belief to predict the receiver's actions, and then uses her own belief to evaluate the receiver's actions.

\paragraph{Smithian Utility of Action}

We define Smithian utility of actions by augmenting the conventional utility of actions defined in \cref{equ:expectation_utility}. There are no subscripts in \cref{equ:expectation_utility}, which implies that the action and belief are from the same agent. \cref{equ:expectation_utility} can apply when one agent evaluates its own action and belief, or when one agent (A) uses theory of mind to take the perspective of another agent (B) to evaluate the agent B's action based on agent B's belief.

With Smithian empathy, the signaler should use her own utility function to evaluate the outcomes (Box 4 in \cref{fig:model}). In many scenarios, the utility functions of the signaler and the receiver are consistent. In some cases, the utility functions may be different; but constrained by Smithian helping, the signaler's utility should always be aligned with the receiver's physical well-being. Here we are using subscripts to represent the source of belief or action, with $Sig$ for signaler and $Rec$ for receiver. We write down the formulation of Smithian utility of action by changing the belief used for action evaluation in \cref{equ:expectation_utility} to the belief of the signaler:
\begin{equation}
    \resizebox{0.91\linewidth}{!}{%
        $\displaystyle{}\E U_{Smith}\left(a_{Rec}| b_{Sig}\right)=\int_{s'\in \mathcal{S}}{U_{Sig}\left(s'\right)\int_{s\in\mathcal{S}}{P\left(s'| s,a_{Rec}\right)P\left(s| b_{Sig}\right)}}.$%
    }
\end{equation}

\paragraph{Smithian Utility of Belief}

The effect of pointing in our study is to change the receiver's belief. Therefore, to evaluate the effect of pointing, we need to define the utility of a belief, especially the receiver's belief without the pointing (Box 1 in \cref{fig:model}) and after the pointing (Box 5 in \cref{fig:model}). It can be derived from the expected utility of actions, as an agent's distribution (Boxes 3 and 7 in \cref{fig:model}) of its action $P(a|b)$ can be predicted from its belief. 

Given $P(a|b)$, we can derive the conventional utility of a belief based on conventional expected utility of action. Let $\mathcal{A}$ be the set of possible actions. The utility of a belief can be defined as:
\begin{equation}
    U\left(b\right)=\int_{a\in\mathcal{A}}{P\left(a| b\right)\E U\left(a|b\right)}.
    \label{equ:belief_utility}
\end{equation}

\cref{equ:belief_utility} can represent an agent's evaluation of its own beliefs. It can also represent an agent A's evaluation of another agent B's beliefs, when A takes B's perspective using Hume's definition of empathy. Using theory of mind, A can predict B's action based on B's belief \cite{wellman2014making}, hence evaluate the belief by integrating out the evaluation of actions.

We can derive Smithian utility of belief by replacing the expected utility of actions in \cref{equ:belief_utility} with the Smithian utility of actions:
\begin{equation}
    \resizebox{0.91\linewidth}{!}{%
        $\displaystyle{}U_{Smith}\left(b_{Rec}| b_{Sig}\right)=\int_{a_{Rec}\in \mathcal{A}_{Rec}}{P\left(a_{Rec}| b_{Rec}\right)\E U_{Smith}\left(a_{Rec}\middle| b_{Sig}\right)}$.
    }
    \label{equ:Smithian_belief}
\end{equation}

Smithian utility of belief represents the signaler's evaluation of the receiver's well-being (Boxes 5 and 8 in \cref{fig:model}), which the signaler tries to improve.

\paragraph{\acf{svi}}

Pointing carries information. In the field of \ac{ai}, the value of information is ``the difference in expected value between the best actions before and after the information is obtained'' \cite{russell2010artificial}, with which an agent can decide how much the information is worth. Here we leverage the insight that the value of information should be calculated as the change in expected utility before and after the information is obtained. \acf{svi} adopts this formalization with two significant differences. First, we calculate the value of information from the perspective of the signaler, not the receiver of the information. Second, the utility we use is the Smithian utility of belief, which is the signaler's estimate of the receiver's well-being. Therefore, \ac{svi} measures the improvement of the signaler's estimate of the receiver's well-being before and after pointing, which is the difference between Box 5 and Box 8 in \cref{fig:model}. In other words, \textbf{\ac{svi} is the utility of pointing}.

To later incorporate our formulation of pointing into \ac{rsa} as the utterance, here we use notation $u$ for pointing. Let $b_{Rec}$ be the receiver's belief before receiving the pointing signal, $b'_{Rec}$ be receiver's belief after receiving the pointing signal:
\begin{equation}
    b'_{Rec}=P_{Rec}(s|u)=P\left(s|b_{Rec},u\right).
\end{equation}
We can write \ac{svi} as:
\begin{equation}
    {\ac{svi}}\left(u|b_{Sig}\right)=U_{Smith}\left(b'_{Rec}|b_{Sig}\right)-U_{Smith}\left(b_{Rec}|b_{Sig}\right).
\end{equation}

\paragraph{Pointing as a Rational Speech Act}

With the utility of pointing clearly defined as \ac{svi}, we can treat pointing as a special type of utterance and model its use and interpretation using \ac{rsa} (Box 9 in \cref{fig:model}). We can write down how a signaler generates pointing $u$:
\begin{equation}
    P_{Sig}\left(u| b_{Sig}\right)\propto \exp\{\alpha\left[\ac{svi}\left(u| b_{Sig}\right)-c\left(u\right)\right]\}.
    \label{equ:RSA_signaler}
\end{equation}
For simplicity of the model, we can set $c(u) = 0$ as the cost of pointing in real world is small.

We still need to adapt \ac{rsa} to consider the entire \ac{pomdp} challenge as the ``context'' of communication. We first outline the \ac{rsa} in language games where the environment is fully observable to both the signaler and receiver. A pragmatic receiver updates its belief upon receiving an utterance $u$ with Bayesian inference \cite{goodman2016pragmatic}
\begin{equation}
    b'_{Rec}\left(s\right)=P_{Rec}\left(s| b_{Rec},u\right)\\
    \; \propto \; P_{Sig}(u|s)P_{Rec}(s|b_{Rec}).
\end{equation}

Then we adapt \ac{rsa} to the case in which neither the signaler nor the receiver can fully observe the environment. In this case, both signaler and receiver maintain a belief based on their observations. Based on existing common ground knowledge, the receiver knows the probability that the signaler has a belief given the physical state $P(b_{Sig}|s)$ \cite{goodman2013knowledge}. The receiver can infer the state of the world with the utterance from the signaler:
\begin{equation}
    P_{Rec}(s|u)\propto \int_{b_{Sig}} P_{Sig}(u|b_{Sig})P(b_{Sig}|s)P(s|b_{Rec}).
    \label{equ:RSA_receiver_full}
\end{equation}

\cref{equ:RSA_receiver_full} can be simplified when the signaler has full knowledge of the world. In this case, the signaler knows a state $s$ for sure without uncertainty. Of course it can still be represented as a belief, a probability distribution with probability 1 on the true state $s^*$, denoted as $b_{Sig}^{s^*}$. The receiver only needs to consider all possible $s$ and their corresponding belief $b_{Sig}^s$. Since the signaler knows the true state, $P(b_{Sig}^{s_i}|s_j)=1$ if $s_i=s_j$, $P(b_{Sig}^{s_i}|s_j)=0$ if $s_i\ne s_j$. The integration in \cref{equ:RSA_receiver_full} has only one nonzero entry for each state. \cref{equ:RSA_receiver_full} reduces to: 
\begin{equation}
    P_{Rec}(s|u)\propto P_{Sig}(u|b_{Sig}^s)P(s|b_{Rec}).
    \label{equ:RSA_receiver_second}
\end{equation}
Assuming $b_{Sig}^s$ is true is equivalent to assuming the signaler knows the true state is $s$. Therefore, using $s$ to denote the true state is $s$, \cref{equ:RSA_receiver_second} further reduces to:
\begin{equation}
    b'_{Rec}\left(s\right) \propto \; P_{Sig}(u|s)P_{Rec}(s|b_{Rec}).
    \label{equ:RSA_receiver_certain}
\end{equation}
In our experiment, we only consider the case that the signaler is certain about the state, so \cref{equ:RSA_receiver_certain} is used in the experiment.

Signal generation and interpretation are modeled through recursive reasoning: For each signal, the signaler estimates the change of receiver's belief after receiving the signal using \cref{equ:RSA_receiver_certain} and selects the signal that maximizes Smithian value of information in \cref{equ:RSA_signaler}. This model of signal generation can then be passed to the next level of receiver for computing the probability of a signal given a world state. This allows the receiver to update its belief as the posterior of the world state given a signal using Bayes' rule. Note that this recursive social reasoning is based on \ac{rsa}. However, in \ac{rsa}, the recursion begins with a literal receiver who infers the reference given an utterance. In our model, the literal receiver is the \ac{pomdp} agent who takes rational actions based on observations. In \cref{equ:Smithian_belief}, to compute $U_{Smith}(b_{Rec}|b_{Sig})$, one needs to compute $P(a_{Rec}|b_{Rec})$ and then integrate out $a$. Here $P(a_{Rec}|b_{Rec})$ is the policy solved by \ac{pomdp}.

\section{Modeling Experiment}

\subsection{Task: Guided Wumpus Hunting}

To highlight the strengths of the Smithian pointing model, we need a single-agent partially observable task as a baseline. We pick a simplified version of the classic \ac{ai} problem: \textit{the Wumpus world} \cite{russell2010artificial}. We augment the task for Smithian pointing by adding an additional helper agent. We call this task the \textit{guided Wumpus hunting game}, directly inspired by the hunting example described in the introduction.

In the guided Wumpus hunting game, a hunter navigates through a set of tiles to shoot a stationary monster, the Wumpus, without knowing its exact location. The Wumpus emits a ``stench'' when it is nearby, which the hunter can smell. The hunter navigates to collect observations, which help him infer the Wumpus's location; however, he cannot go near the Wumpus's location. He has one arrow that he can shoot to kill the Wumpus from a distance, based on where he believes the Wumpus to be. In addition, for a second agent---the guide---the environment is fully observable. However, her communication to the hunter is minimal: she can only decide whether to point to a stench after the hunter observes it, without specifying what she hopes the hunter to do with that stench. This setting is directly inspired by the ``broken stick'' example, to capture the overloadedness and indirectedness of pointing. Here, although the state space is smaller than the classic Wumpus world, the inference is in fact computationally more expensive as the POMDP solver is recursively called by \ac{rsa}. 

\begin{figure}[t!]
    \centering
    \includegraphics[width=0.6\linewidth]{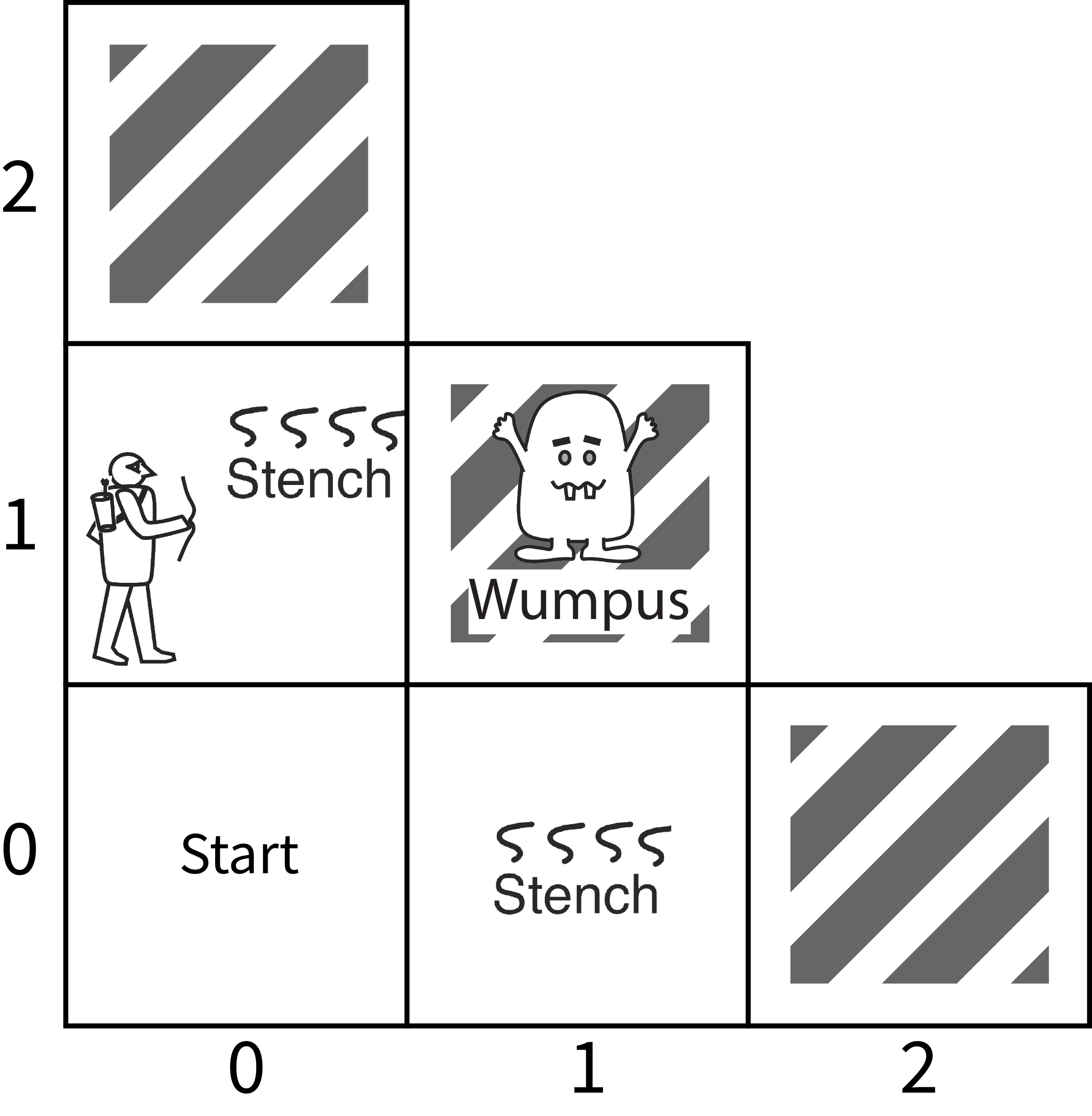}
    \caption{\textbf{The environment of the guided Wumpus hunting game.} Wumpus can only show up in one of three shaded tiles. Starting from (0, 0), the hunter tries to infer Wumpus's location from the stench and shoot it.} 
    \label{fig:env}
\end{figure}

\subsection{Game}

The guided Wumpus hunting game has an environment shown in \cref{fig:env}. An agent plays this game with a \ac{pomdp} solver. Components of the \ac{pomdp} formulation are introduced below. 

\paragraph{State Space}
The Wumpus can show up in one of the three possible locations: (0, 2), (1, 1), and (2, 0). Hunters will always start from (0, 0) and explore 3 possible locations for collecting observations: (0, 0), (0, 1), and (1, 0).

\paragraph{Action Space}
The hunter can choose from 4 possible actions: move vertically, move horizontally, shoot to the upper tile, or shoot to the right tile, which are the only possible directions given the map.

\paragraph{Transition Function}
The outcomes of the hunter's actions are deterministic. At each step, he can decide to move or shoot. With moving, he will always move one tile in the direction of the action. If the hunter moves outside of the map, he will return to (0, 0). For shooting, the arrow will hit or miss the Wumpus depending on whether the Wumpus is in the shoot direction. The game ends after shooting. 
 
\paragraph{Reward Function}
For moving one step, there is an action cost, which is manipulated in our experiment. When shooting, the reward is set to 100 for hitting the Wumpus, -100 for missing. 

\paragraph{Observation Space and Observation Function}
There are only two possible observations: a stench or nothing. The observation function is stochastic. If a tile is near the Wumpus, the hunter's probability of observing a stench in that tile is 0.85. If a tile is not nearby the Wumpus, the probability of observing a stench is 0.15. In the classic Wumpus world, the observation is deterministic; however, we add stochasticity to make the belief inference more interesting, as in other \ac{pomdp} tasks. We do not place pointing in the observation space as it is modeled as an communicative act, instead of an observation of the physical world \cite{kaelbling1998planning}.

\subsection{Experiment}

We designed an experiment with the guided Wumpus hunting game to compare the models of individually perceived observations and jointly perceived observations. We manipulate the hunter's model of interpreting the observation and the moving cost in the game.

We use the \ac{pomdp} as the baseline model of a hunter who individually perceives the observation. In this condition, the hunter, as a receiver, will completely ignore the pointing signal sent by the guide.

We use Smithian pointing as the model of a hunter who pragmatically perceives the observation pointed to by the guide. The guide will use the Smithian pointing model to generate signals. Since the purpose of pointing is to help the receiver, we predict that hunters who use the Smithian pointing model will outperform those who use \acp{pomdp}.

It is possible that this predicted improvement in performance is simply caused by the amount of information provided by pointing but not the pragmatic inference process. To test this possibility, we add a third condition. In this condition, the hunter uses \acp{pomdp} to individually perceive observations, but when receiving a pointing signal, he receives an additional observation to which the pointing is directed. We call this condition \acp{pomdp} with``double observations''.

We also manipulate the cost of each step in the environment. The purpose of the guide pointing is to help the hunter achieve higher performance, but when the cost of moving is too high or low, help becomes unnecessary. If the moving cost is too high, for example -9, the hunter will shoot in a rush without considering the effect of observations. If the moving cost is too low, for example -1, the hunter will move around the environment more actively to collect observations no matter whether the guide points or not. The effect of communication as a function of moving cost may not be linear; therefore, in our experiment, we test 100 trials under each model using a moving cost of -9, -7, -5, -3, and -1. We expect to see the effect of pointing when the moving cost is moderate, but not when it is too high (-9) or too low (-1).

\section{Result}

\begin{figure}[t!]
    \centering
    \includegraphics[width=0.9\linewidth]{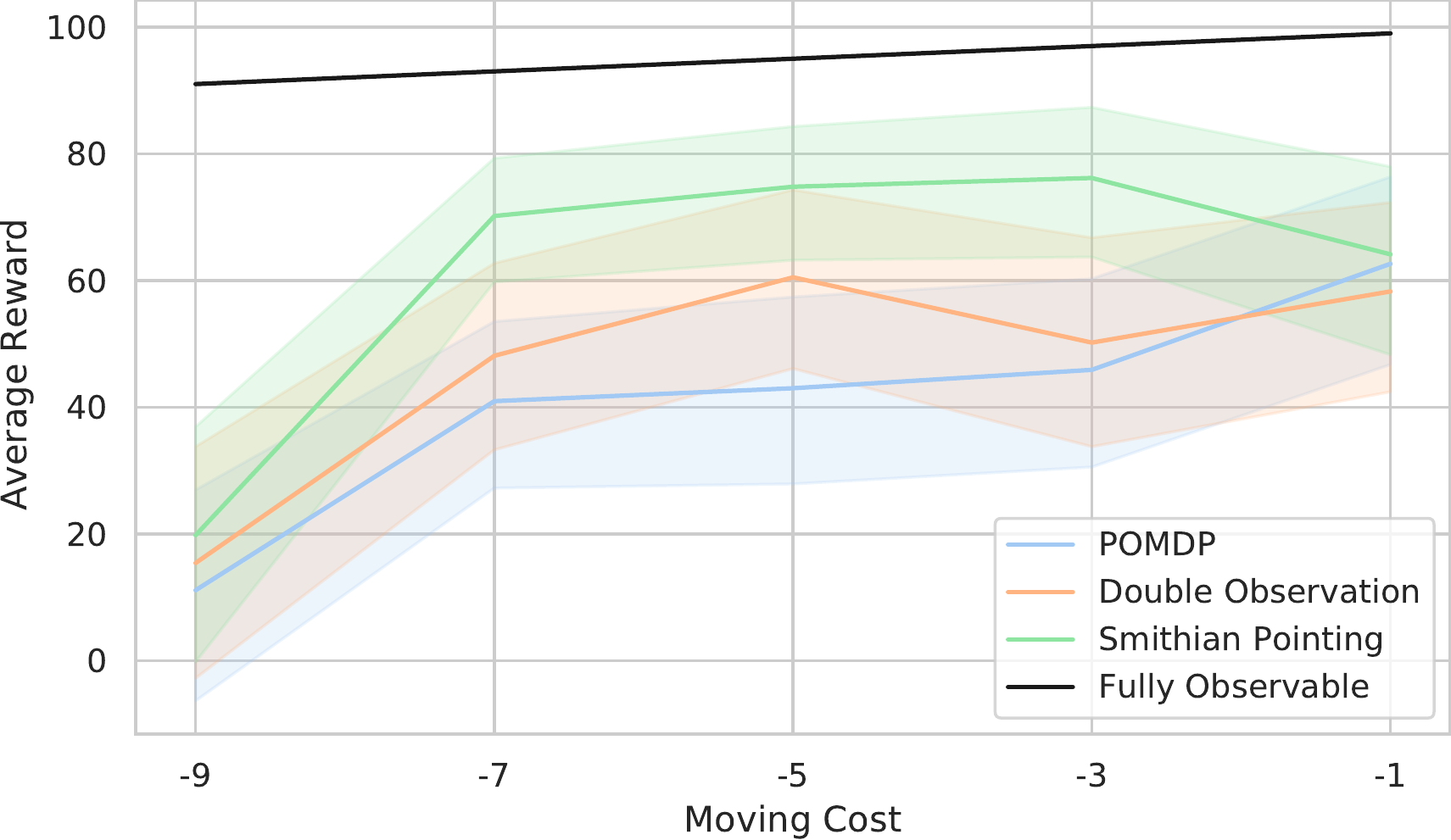}
    \caption{\textbf{Experimental results.} Black line denotes the ideal performance upper bound if the game was fully observable. Shaded areas represent $95\%$ bootstrap confidence interval.}
    \label{fig:result}
\end{figure}

The average reward across trials for each model under various moving cost is depicted in \cref{fig:result}. Overall, the proposed Smithian pointing model achieves better performance compared to the classic \ac{pomdp} model or the \ac{pomdp} with double observations. The main effect of model type is significant ($F (2, 1485) = 9.602$, $p < 0.001$), and the main effect of moving cost is also significant ($F (4, 1485) = 19.535$, $p < 0.001$). However, the interaction between models and moving costs is not significant ($F (8, 1485) = 1.035$, $p = 0.407$). A post-hoc test with bonferroni correction shows that the Smithian pointing model achieves higher performance than ``Double observation'' condition ($F (1, 998) = 8.875$, $p = 0.009$). As hypothesized, our experiment also shows that the advantage stemmed from the pragmatic inference of pointing disappears when the task is too hard/easy for the guide to help. Specifically, when the moving cost is -1 or -9, the effect of model type is not significant ($F (2, 297) = 0.163$, $p = 0.850$; $F (2, 297) = 0.228$, $p = 0.796$). Taken together, these results demonstrate that pointing is relevant only when the signaler could offer help. Our computational model captures this relevance \cite{sperber1986relevance} and highlights how joint perception can be more powerful than individual perception of the same observation, demonstrated by the improved hunting performance.

\section{Discussion}

In this paper, we devise a computational model for pointing by defining the \acf{svi}, applying it to \ac{rsa} to define the utility of pointing. In an example task, our pointing model shows a significant performance improvement compared to a single-agent \ac{pomdp} or a single-agent \ac{pomdp} with ``double observation''. This improvement indicates that the advantage of pointing does not come from providing a new observation for individual perception. Instead, it comes from the pragmatic inference of how the jointly perceived observation is relevant to the task. Supporting the argument, our experiment also shows that the advantage of the Smithian pointing model works the best only when the receiver is in a position to be helped.

Our results suggest that Smithian coordination of beliefs is necessary for modeling pointing. Seemingly simple, pointing requires the intelligent social capacity of mind coordination and altruistic motivation of helping. As a type of sophisticated social cognition, pointing involves identifying receiver's belief to predict action, evaluating actions with signaler's own belief, generating pointing to help change receiver's mind, and interpreting pointing with the assumption that the signaler is trying to help. The possible lack of some of these components, especially the cooperative motivation in pointing generation and interpretation, may explain the rarity of pointing in wild non-human primates. Indeed, the ability of generating and interpreting pointing is a milestone in human unique communication \cite{tomasello2010origins}.   

Due to the complexity of recursive reasoning of \ac{rsa}, \ac{pomdp} is solved multiple times in the recursion, which is computationally expensive and limits the current experiment setting to have a small state space. In follow up studies, we can use a faster \ac{pomdp} solver or approximate the social recursion using computationally cheaper approaches \cite{kaelbling2013integrated}. In this way, we can expand our model to more complex tasks with larger state space and more observations. As many insights of this paper directly come from studies of child development, we hope this work will foster further interdisciplinary studies between developmental psychology and \ac{ai}.

\section{Acknowledgments}

This research was supported by DARPA PA 19-03-01 and ONR MURI project N00014-16-1-2007 to TG.

\bibliographystyle{apacite}

\setlength{\bibleftmargin}{.125in}
\setlength{\bibindent}{-\bibleftmargin}
\bibliography{reference}

\end{document}

%% file: config.tex
\usepackage{graphicx}
\usepackage{amsmath,amssymb,amsthm,mathabx}
\usepackage{acronym}
\usepackage{balance}
\usepackage{xspace}
\usepackage{setspace}
\usepackage[skip=3pt,font=small]{subcaption}
\usepackage[capitalise]{cleveref}

\makeatletter
\DeclareRobustCommand\onedot{\futurelet\@let@token\@onedot}
\def\@onedot{\ifx\@let@token.\else.\null\fi\xspace}

\makeatother

\DeclareMathOperator{\E}{\mathbb{E}}

\makeatletter
\renewcommand{\paragraph}{%
  \@startsection{paragraph}{4}%
  {\z@}{0ex \@plus 0ex \@minus 0ex}{-1em}%
  {\hskip\parindent\normalfont\normalsize\bfseries}%
}
\makeatother

\graphicspath{{figures/}}

\crefname{algocf}{Alg.}{Algs.}
\Crefname{algocf}{Algrithm}{Algrithm}

\acrodef{ai}[AI]{artificial intelligence}
\acrodef{rsa}[RSA]{rational speech act}
\acrodef{pomdp}[POMDP]{partially observable Markov decision process}
\acrodef{svi}[SVI]{Smithian Value of Information}

%% file: main.bbl
\begin{thebibliography}{}

\bibitem [\protect \citeauthoryear {%
Baker%
, Saxe%
\BCBL {}\ \BBA {} Tenenbaum%
}{%
Baker%
\ \protect \BOthers {.}}{%
{\protect \APACyear {2011}}%
}]{%
baker2011bayesian}
\APACinsertmetastar {%
baker2011bayesian}%
\begin{APACrefauthors}%
Baker, C.%
, Saxe, R.%
\BCBL {}\ \BBA {} Tenenbaum, J.%
\end{APACrefauthors}%
\unskip\
\newblock
\APACrefYearMonthDay{2011}{}{}.
\newblock
{\BBOQ}\APACrefatitle {Bayesian theory of mind: Modeling joint belief-desire
  attribution} {Bayesian theory of mind: Modeling joint belief-desire
  attribution}.{\BBCQ}
\newblock
\BIn{} \APACrefbtitle {Proceedings of the Annual Meeting of the Cognitive
  Science Society} {Proceedings of the annual meeting of the cognitive science
  society}\ (\BPGS\ 2469--2474).
\PrintBackRefs{\CurrentBib}

\bibitem [\protect \citeauthoryear {%
Butterworth%
, Simion%
\BCBL {}\ \protect \BOthers {.}}{%
Butterworth%
\ \protect \BOthers {.}}{%
{\protect \APACyear {2013}}%
}]{%
butterworth2013development}
\APACinsertmetastar {%
butterworth2013development}%
\begin{APACrefauthors}%
Butterworth, G.%
, Simion, F.%
\BCBL {}\ \BOthersPeriod {.}\end{APACrefauthors}%
\unskip\
\newblock
\APACrefYear{2013}.
\newblock
\APACrefbtitle {The Development Of Sensory, Motor And Cognitive Capacities In
  Early Infancy: From Sensation To Cognition} {The development of sensory,
  motor and cognitive capacities in early infancy: From sensation to
  cognition}.
\newblock
\APACaddressPublisher{}{Psychology Press}.
\PrintBackRefs{\CurrentBib}

\bibitem [\protect \citeauthoryear {%
Frank%
\ \BBA {} Goodman%
}{%
Frank%
\ \BBA {} Goodman%
}{%
{\protect \APACyear {2012}}%
}]{%
frank2012predicting}
\APACinsertmetastar {%
frank2012predicting}%
\begin{APACrefauthors}%
Frank, M\BPBI C.%
\BCBT {}\ \BBA {} Goodman, N\BPBI D.%
\end{APACrefauthors}%
\unskip\
\newblock
\APACrefYearMonthDay{2012}{}{}.
\newblock
{\BBOQ}\APACrefatitle {Predicting pragmatic reasoning in language games}
  {Predicting pragmatic reasoning in language games}.{\BBCQ}
\newblock
\APACjournalVolNumPages{Science}{336}{6084}{998--998}.
\PrintBackRefs{\CurrentBib}

\bibitem [\protect \citeauthoryear {%
Goodman%
\ \BBA {} Frank%
}{%
Goodman%
\ \BBA {} Frank%
}{%
{\protect \APACyear {2016}}%
}]{%
goodman2016pragmatic}
\APACinsertmetastar {%
goodman2016pragmatic}%
\begin{APACrefauthors}%
Goodman, N\BPBI D.%
\BCBT {}\ \BBA {} Frank, M\BPBI C.%
\end{APACrefauthors}%
\unskip\
\newblock
\APACrefYearMonthDay{2016}{}{}.
\newblock
{\BBOQ}\APACrefatitle {Pragmatic language interpretation as probabilistic
  inference} {Pragmatic language interpretation as probabilistic
  inference}.{\BBCQ}
\newblock
\APACjournalVolNumPages{Trends in Cognitive Sciences}{20}{11}{818--829}.
\PrintBackRefs{\CurrentBib}

\bibitem [\protect \citeauthoryear {%
Goodman%
\ \BBA {} Stuhlm{\"u}ller%
}{%
Goodman%
\ \BBA {} Stuhlm{\"u}ller%
}{%
{\protect \APACyear {2013}}%
}]{%
goodman2013knowledge}
\APACinsertmetastar {%
goodman2013knowledge}%
\begin{APACrefauthors}%
Goodman, N\BPBI D.%
\BCBT {}\ \BBA {} Stuhlm{\"u}ller, A.%
\end{APACrefauthors}%
\unskip\
\newblock
\APACrefYearMonthDay{2013}{}{}.
\newblock
{\BBOQ}\APACrefatitle {Knowledge and implicature: Modeling language
  understanding as social cognition} {Knowledge and implicature: Modeling
  language understanding as social cognition}.{\BBCQ}
\newblock
\APACjournalVolNumPages{Topics in Cognitive Science}{5}{1}{173--184}.
\PrintBackRefs{\CurrentBib}

\bibitem [\protect \citeauthoryear {%
Grice%
}{%
Grice%
}{%
{\protect \APACyear {1975}}%
}]{%
grice1975logic}
\APACinsertmetastar {%
grice1975logic}%
\begin{APACrefauthors}%
Grice, H\BPBI P.%
\end{APACrefauthors}%
\unskip\
\newblock
\APACrefYearMonthDay{1975}{}{}.
\newblock
{\BBOQ}\APACrefatitle {Logic and conversation} {Logic and conversation}.{\BBCQ}
\newblock
\BIn{} P.~Cole\ \BBA {} J\BPBI L.~Morgan\ (\BEDS), \APACrefbtitle {Speech Acts}
  {Speech acts}\ (\BPGS\ 41--58).
\newblock
\APACaddressPublisher{}{Brill}.
\PrintBackRefs{\CurrentBib}

\bibitem [\protect \citeauthoryear {%
Hepach%
, Benziad%
\BCBL {}\ \BBA {} Tomasello%
}{%
Hepach%
\ \protect \BOthers {.}}{%
{\protect \APACyear {2020}}%
}]{%
hepach2020chimpanzees}
\APACinsertmetastar {%
hepach2020chimpanzees}%
\begin{APACrefauthors}%
Hepach, R.%
, Benziad, L.%
\BCBL {}\ \BBA {} Tomasello, M.%
\end{APACrefauthors}%
\unskip\
\newblock
\APACrefYearMonthDay{2020}{}{}.
\newblock
{\BBOQ}\APACrefatitle {Chimpanzees help others with what they want; children
  help them with what they need} {Chimpanzees help others with what they want;
  children help them with what they need}.{\BBCQ}
\newblock
\APACjournalVolNumPages{Developmental Science}{23}{3}{Article e12922}.
\PrintBackRefs{\CurrentBib}

\bibitem [\protect \citeauthoryear {%
Hume%
}{%
Hume%
}{%
{\protect \APACyear {1751/2018}}%
}]{%
hume2018enquiry}
\APACinsertmetastar {%
hume2018enquiry}%
\begin{APACrefauthors}%
Hume, D.%
\end{APACrefauthors}%
\unskip\
\newblock
\APACrefYear{2018}.
\newblock
\APACrefbtitle {An enquiry concerning the principles of morals} {An enquiry
  concerning the principles of morals}.
\newblock
\APACaddressPublisher{}{Yale University Press}.
\newblock
\APACorigyearnote{1751}{}
\PrintBackRefs{\CurrentBib}

\bibitem [\protect \citeauthoryear {%
Kaelbling%
, Littman%
\BCBL {}\ \BBA {} Cassandra%
}{%
Kaelbling%
\ \protect \BOthers {.}}{%
{\protect \APACyear {1998}}%
}]{%
kaelbling1998planning}
\APACinsertmetastar {%
kaelbling1998planning}%
\begin{APACrefauthors}%
Kaelbling, L\BPBI P.%
, Littman, M\BPBI L.%
\BCBL {}\ \BBA {} Cassandra, A\BPBI R.%
\end{APACrefauthors}%
\unskip\
\newblock
\APACrefYearMonthDay{1998}{}{}.
\newblock
{\BBOQ}\APACrefatitle {Planning and acting in partially observable stochastic
  domains} {Planning and acting in partially observable stochastic
  domains}.{\BBCQ}
\newblock
\APACjournalVolNumPages{Artificial Intelligence}{101}{1-2}{99--134}.
\PrintBackRefs{\CurrentBib}

\bibitem [\protect \citeauthoryear {%
Kaelbling%
\ \BBA {} Lozano-P{\'e}rez%
}{%
Kaelbling%
\ \BBA {} Lozano-P{\'e}rez%
}{%
{\protect \APACyear {2013}}%
}]{%
kaelbling2013integrated}
\APACinsertmetastar {%
kaelbling2013integrated}%
\begin{APACrefauthors}%
Kaelbling, L\BPBI P.%
\BCBT {}\ \BBA {} Lozano-P{\'e}rez, T.%
\end{APACrefauthors}%
\unskip\
\newblock
\APACrefYearMonthDay{2013}{}{}.
\newblock
{\BBOQ}\APACrefatitle {Integrated task and motion planning in belief space}
  {Integrated task and motion planning in belief space}.{\BBCQ}
\newblock
\APACjournalVolNumPages{The International Journal of Robotics
  Research}{32}{9-10}{1194--1227}.
\PrintBackRefs{\CurrentBib}

\bibitem [\protect \citeauthoryear {%
Knill%
\ \BBA {} Richards%
}{%
Knill%
\ \BBA {} Richards%
}{%
{\protect \APACyear {1996}}%
}]{%
knill1996perception}
\APACinsertmetastar {%
knill1996perception}%
\begin{APACrefauthors}%
Knill, D\BPBI C.%
\BCBT {}\ \BBA {} Richards, W.%
\end{APACrefauthors}%
\unskip\
\newblock
\APACrefYear{1996}.
\newblock
\APACrefbtitle {Perception as Bayesian inference} {Perception as bayesian
  inference}.
\newblock
\APACaddressPublisher{}{Cambridge University Press}.
\PrintBackRefs{\CurrentBib}

\bibitem [\protect \citeauthoryear {%
Leavens%
\ \BBA {} Hopkins%
}{%
Leavens%
\ \BBA {} Hopkins%
}{%
{\protect \APACyear {1998}}%
}]{%
leavens1998intentional}
\APACinsertmetastar {%
leavens1998intentional}%
\begin{APACrefauthors}%
Leavens, D\BPBI A.%
\BCBT {}\ \BBA {} Hopkins, W\BPBI D.%
\end{APACrefauthors}%
\unskip\
\newblock
\APACrefYearMonthDay{1998}{}{}.
\newblock
{\BBOQ}\APACrefatitle {Intentional communication by chimpanzees: a
  cross-sectional study of the use of referential gestures.} {Intentional
  communication by chimpanzees: a cross-sectional study of the use of
  referential gestures.}{\BBCQ}
\newblock
\APACjournalVolNumPages{Developmental Psychology}{34}{5}{813}.
\PrintBackRefs{\CurrentBib}

\bibitem [\protect \citeauthoryear {%
Liebal%
, Carpenter%
\BCBL {}\ \BBA {} Tomasello%
}{%
Liebal%
\ \protect \BOthers {.}}{%
{\protect \APACyear {2011}}%
}]{%
liebal2011young}
\APACinsertmetastar {%
liebal2011young}%
\begin{APACrefauthors}%
Liebal, K.%
, Carpenter, M.%
\BCBL {}\ \BBA {} Tomasello, M.%
\end{APACrefauthors}%
\unskip\
\newblock
\APACrefYearMonthDay{2011}{}{}.
\newblock
{\BBOQ}\APACrefatitle {Young children's understanding of markedness in
  non-verbal communication} {Young children's understanding of markedness in
  non-verbal communication}.{\BBCQ}
\newblock
\APACjournalVolNumPages{Journal of Child Language}{38}{4}{888}.
\PrintBackRefs{\CurrentBib}

\bibitem [\protect \citeauthoryear {%
Liszkowski%
, Carpenter%
, Striano%
\BCBL {}\ \BBA {} Tomasello%
}{%
Liszkowski%
\ \protect \BOthers {.}}{%
{\protect \APACyear {2006}}%
}]{%
liszkowski2006twelveand}
\APACinsertmetastar {%
liszkowski2006twelveand}%
\begin{APACrefauthors}%
Liszkowski, U.%
, Carpenter, M.%
, Striano, T.%
\BCBL {}\ \BBA {} Tomasello, M.%
\end{APACrefauthors}%
\unskip\
\newblock
\APACrefYearMonthDay{2006}{}{}.
\newblock
{\BBOQ}\APACrefatitle {12-and 18-month-olds point to provide information for
  others} {12-and 18-month-olds point to provide information for
  others}.{\BBCQ}
\newblock
\APACjournalVolNumPages{Journal of cognition and development}{7}{2}{173--187}.
\PrintBackRefs{\CurrentBib}

\bibitem [\protect \citeauthoryear {%
Martin%
, Lin%
\BCBL {}\ \BBA {} Olson%
}{%
Martin%
\ \protect \BOthers {.}}{%
{\protect \APACyear {2016}}%
}]{%
martin2016you}
\APACinsertmetastar {%
martin2016you}%
\begin{APACrefauthors}%
Martin, A.%
, Lin, K.%
\BCBL {}\ \BBA {} Olson, K\BPBI R.%
\end{APACrefauthors}%
\unskip\
\newblock
\APACrefYearMonthDay{2016}{}{}.
\newblock
{\BBOQ}\APACrefatitle {What you want versus what's good for you: Paternalistic
  motivation in children's helping behavior} {What you want versus what's good
  for you: Paternalistic motivation in children's helping behavior}.{\BBCQ}
\newblock
\APACjournalVolNumPages{Child development}{87}{6}{1739--1746}.
\PrintBackRefs{\CurrentBib}

\bibitem [\protect \citeauthoryear {%
Pineau%
, Gordon%
\BCBL {}\ \BBA {} Thrun%
}{%
Pineau%
\ \protect \BOthers {.}}{%
{\protect \APACyear {2003}}%
}]{%
pineau2003point}
\APACinsertmetastar {%
pineau2003point}%
\begin{APACrefauthors}%
Pineau, J.%
, Gordon, G.%
\BCBL {}\ \BBA {} Thrun, S.%
\end{APACrefauthors}%
\unskip\
\newblock
\APACrefYearMonthDay{2003}{}{}.
\newblock
{\BBOQ}\APACrefatitle {Point-based value iteration: An anytime algorithm for
  POMDPs} {Point-based value iteration: An anytime algorithm for
  pomdps}.{\BBCQ}
\newblock
\BIn{} \APACrefbtitle {Proceedings of International Joint Conference on
  Artificial Intelligence} {Proceedings of international joint conference on
  artificial intelligence}\ (\BPGS\ 1025--1032).
\PrintBackRefs{\CurrentBib}

\bibitem [\protect \citeauthoryear {%
Russell%
, Norvig%
\BCBL {}\ \BBA {} Davis%
}{%
Russell%
\ \protect \BOthers {.}}{%
{\protect \APACyear {2010}}%
}]{%
russell2010artificial}
\APACinsertmetastar {%
russell2010artificial}%
\begin{APACrefauthors}%
Russell, S\BPBI J.%
, Norvig, P.%
\BCBL {}\ \BBA {} Davis, E.%
\end{APACrefauthors}%
\unskip\
\newblock
\APACrefYear{2010}.
\newblock
\APACrefbtitle {Artificial intelligence: a modern approach} {Artificial
  intelligence: a modern approach}\ (\PrintOrdinal{3}\ \BEd).
\newblock
\APACaddressPublisher{}{Prentice Hall}.
\PrintBackRefs{\CurrentBib}

\bibitem [\protect \citeauthoryear {%
Smith%
}{%
Smith%
}{%
{\protect \APACyear {1759/2010}}%
}]{%
smith2010theory}
\APACinsertmetastar {%
smith2010theory}%
\begin{APACrefauthors}%
Smith, A.%
\end{APACrefauthors}%
\unskip\
\newblock
\APACrefYear{2010}.
\newblock
\APACrefbtitle {The theory of moral sentiments} {The theory of moral
  sentiments}.
\newblock
\APACaddressPublisher{}{Penguin}.
\newblock
\APACorigyearnote{1759}{}
\PrintBackRefs{\CurrentBib}

\bibitem [\protect \citeauthoryear {%
Sperber%
\ \BBA {} Wilson%
}{%
Sperber%
\ \BBA {} Wilson%
}{%
{\protect \APACyear {1986}}%
}]{%
sperber1986relevance}
\APACinsertmetastar {%
sperber1986relevance}%
\begin{APACrefauthors}%
Sperber, D.%
\BCBT {}\ \BBA {} Wilson, D.%
\end{APACrefauthors}%
\unskip\
\newblock
\APACrefYear{1986}.
\newblock
\APACrefbtitle {Relevance: Communication and cognition} {Relevance:
  Communication and cognition}.
\newblock
\APACaddressPublisher{}{Harvard University Press Cambridge, MA}.
\PrintBackRefs{\CurrentBib}

\bibitem [\protect \citeauthoryear {%
Tomasello%
}{%
Tomasello%
}{%
{\protect \APACyear {2010}}%
}]{%
tomasello2010origins}
\APACinsertmetastar {%
tomasello2010origins}%
\begin{APACrefauthors}%
Tomasello, M.%
\end{APACrefauthors}%
\unskip\
\newblock
\APACrefYear{2010}.
\newblock
\APACrefbtitle {Origins of human communication} {Origins of human
  communication}.
\newblock
\APACaddressPublisher{}{MIT press}.
\PrintBackRefs{\CurrentBib}

\bibitem [\protect \citeauthoryear {%
Ullman%
\ \protect \BOthers {.}}{%
Ullman%
\ \protect \BOthers {.}}{%
{\protect \APACyear {2009}}%
}]{%
ullman2009help}
\APACinsertmetastar {%
ullman2009help}%
\begin{APACrefauthors}%
Ullman, T\BPBI D.%
, Baker, C\BPBI L.%
, Macindoe, O.%
, Evans, O.%
, Goodman, N\BPBI D.%
\BCBL {}\ \BBA {} Tenenbaum, J\BPBI B.%
\end{APACrefauthors}%
\unskip\
\newblock
\APACrefYearMonthDay{2009}{}{}.
\newblock
{\BBOQ}\APACrefatitle {Help or hinder: Bayesian models of social goal
  inference} {Help or hinder: Bayesian models of social goal inference}.{\BBCQ}
\newblock
\BIn{} \APACrefbtitle {Proceedings of the 22nd International Conference on
  Neural Information Processing Systems} {Proceedings of the 22nd international
  conference on neural information processing systems}\ (\BPGS\ 1874--1882).
\PrintBackRefs{\CurrentBib}

\bibitem [\protect \citeauthoryear {%
Van~der Goot%
, Tomasello%
\BCBL {}\ \BBA {} Liszkowski%
}{%
Van~der Goot%
\ \protect \BOthers {.}}{%
{\protect \APACyear {2014}}%
}]{%
van2014differences}
\APACinsertmetastar {%
van2014differences}%
\begin{APACrefauthors}%
Van~der Goot, M\BPBI H.%
, Tomasello, M.%
\BCBL {}\ \BBA {} Liszkowski, U.%
\end{APACrefauthors}%
\unskip\
\newblock
\APACrefYearMonthDay{2014}{}{}.
\newblock
{\BBOQ}\APACrefatitle {Differences in the nonverbal requests of great apes and
  human infants} {Differences in the nonverbal requests of great apes and human
  infants}.{\BBCQ}
\newblock
\APACjournalVolNumPages{Child Development}{85}{2}{444--455}.
\PrintBackRefs{\CurrentBib}

\bibitem [\protect \citeauthoryear {%
Wellman%
}{%
Wellman%
}{%
{\protect \APACyear {2014}}%
}]{%
wellman2014making}
\APACinsertmetastar {%
wellman2014making}%
\begin{APACrefauthors}%
Wellman, H\BPBI M.%
\end{APACrefauthors}%
\unskip\
\newblock
\APACrefYear{2014}.
\newblock
\APACrefbtitle {Making minds: How theory of mind develops} {Making minds: How
  theory of mind develops}.
\newblock
\APACaddressPublisher{}{Oxford University Press}.
\PrintBackRefs{\CurrentBib}

\bibitem [\protect \citeauthoryear {%
Wimmer%
\ \BBA {} Perner%
}{%
Wimmer%
\ \BBA {} Perner%
}{%
{\protect \APACyear {1983}}%
}]{%
wimmer1983beliefs}
\APACinsertmetastar {%
wimmer1983beliefs}%
\begin{APACrefauthors}%
Wimmer, H.%
\BCBT {}\ \BBA {} Perner, J.%
\end{APACrefauthors}%
\unskip\
\newblock
\APACrefYearMonthDay{1983}{}{}.
\newblock
{\BBOQ}\APACrefatitle {Beliefs about beliefs: Representation and constraining
  function of wrong beliefs in young children's understanding of deception}
  {Beliefs about beliefs: Representation and constraining function of wrong
  beliefs in young children's understanding of deception}.{\BBCQ}
\newblock
\APACjournalVolNumPages{Cognition}{13}{1}{103--128}.
\PrintBackRefs{\CurrentBib}

\bibitem [\protect \citeauthoryear {%
Wittgenstein%
\ \BBA {} Anscombe%
}{%
Wittgenstein%
\ \BBA {} Anscombe%
}{%
{\protect \APACyear {1953/2001}}%
}]{%
wittgenstein2001philosophical}
\APACinsertmetastar {%
wittgenstein2001philosophical}%
\begin{APACrefauthors}%
Wittgenstein, L.%
\BCBT {}\ \BBA {} Anscombe, G.%
\end{APACrefauthors}%
\unskip\
\newblock
\APACrefYear{2001}.
\newblock
\APACrefbtitle {Philosophical investigations : the german text, with a revised
  english translation} {Philosophical investigations : the german text, with a
  revised english translation}\ (\PrintOrdinal{3}\ \BEd).
\newblock
\APACaddressPublisher{}{Blackwell}.
\newblock
\APACorigyearnote{1953}{}
\PrintBackRefs{\CurrentBib}

\end{thebibliography}
